\documentclass[cameraready]{Interspeech}

\usepackage{CJKutf8} %
\usepackage{xcolor} %

\usepackage{booktabs}
\usepackage{multirow}
\usepackage{tabularx}  

\usepackage{amsmath}

\usepackage{xcolor}
\usepackage{mdframed}

\usepackage{tikz}
\usepackage[export]{adjustbox}               %
\usepackage[skip=1ex, belowskip=2ex]{subcaption}
\usepackage{float}

\usepackage{xcolor}

\definecolor{eqA}{HTML}{D7263D}
\definecolor{eqB}{HTML}{1B998B}
\definecolor{eqC}{HTML}{2E86AB}
\definecolor{eqD}{HTML}{F46036}
\definecolor{eqE}{HTML}{6A4C93}
\definecolor{eqF}{HTML}{3A86FF}

\mdfdefinestyle{GreenBox}{
  linecolor=black,
  linewidth=0.5pt,
  backgroundcolor=gray!10,
  innermargin=5pt,
  outermargin=5pt,
  innertopmargin=5pt,
  innerbottommargin=5pt
}

\title{Languages in Whisper-Style Speech Encoders \\Align Both Phonetically and Semantically}

\author[affiliation={1,2}]{Ryan Soh-Eun}{Shim}
\author[affiliation={3}]{Domenico}{De Cristofaro}
\author[affiliation={4}]{Martin Chengzhi}{Hu}
\author[affiliation={3}]{Alessandro}{Vietti}
\author[affiliation={1,2}]{Barbara}{Plank}

\address{
    $^1$ LMU Munich,
    $^2$ Munich Center for Machine Learning,
    $^3$ Free University of Bozen-Bolzano,
    $^4$ Technical University of Munich 
}

\email{s.shim@lmu.de}


\usepackage{comment}

\begin{document}

\maketitle

\begin{abstract}

Cross-lingual alignment in pretrained language models enables knowledge transfer across languages. Similar alignment has been reported in Whisper-style speech encoders, based on spoken translation retrieval using representational similarity. However, prior work does not control for phonetic overlap between equivalent utterances, which may artificially support retrieval. We conduct pronunciation-controlled experiments to test whether cross-lingual alignment arises from semantic rather than phonetic similarity. Results show that spoken translation retrieval remains strongly above chance without phonetic cues in the final layers of encoders trained with a speech translation objective, most clearly for models additionally trained on translation. We further test early-exiting the encoder to induce representations we hypothesize to be less tied to language-specific semantics. These experiments indeed reveal performance gains in automatic speech recognition on low-resource languages unseen during training. %

\end{abstract}

\section{Introduction}

In speech, a growing body of work has shown speech foundation models to exhibit emergent multilingual capabilities \cite{peng23d_interspeech, whisper-cs}, implying the existence of cross-lingual alignment in such models. Prior work probes such alignment through spoken translation retrieval \cite{abdullah24_interspeech, ma2025crosslingualtransferlearningspeech}, where \cite{ma2025crosslingualtransferlearningspeech} find Whisper's encoder to have \textit{an accuracy of up to 80\% on the FLEURS dataset for language pairs such as English and French}. However, a possible confounding factor in using speech retrieval as a proxy for measuring cross-lingual alignment is the existence of \textit{pronunciation}-level cues such as cognates, loanwords, and proper nouns, that could serve as shortcuts even in the absence of a shared semantic space for retrieving semantically equivalent utterances in a different language (\autoref{tab:pron-shortcuts}). As such, in this paper we ask: \emph{to what degree does spoken translation retrieval actually rely on semantic features?} To answer this question, we explicitly control for such pronunciation-level cues by constructing a challenge set without pronunciation-level cues between typologically-distant languages, with the goal of understanding whether and to what extent such shortcuts impact the cross-lingual alignment in speech. In addition, to observe whether such an alignment is induced by the speech translation objective or whether it can arise through the next token prediction objective in ASR alone, we perform a controlled study on comparable Whisper-style speech encoders trained with and without the speech translation objective, where we find the latter to observe weak but non-trivial spoken translation retrieval capabilities. Finally, leveraging our findings above, we perform early exiting experiments in the encoder layers to induce representations less tied to language-specific semantics, yielding performance gains for speech recognition
on low-resource languages unseen during training. The contributions of our paper are as follows:

\begin{table}[tbp]
\centering

\begin{CJK*}{UTF8}{gbsn}
\begin{mdframed}[style=GreenBox]
  \footnotesize
  \begin{tabularx}{\textwidth}{@{}lX@{}}
    English: & You can \textbf{\textcolor{red}{use}} a \textbf{\textcolor{orange}{boda-boda}} to get around \textbf{\textcolor{purple}{Goma}}. The \textbf{\textcolor{pink}{normal price}} is \(\sim\)\textbf{\textcolor{blue}{500 Congolese Francs}} for a short ride. \\
    Italian: & È possibile \textbf{\textcolor{red}{usare}} un \textbf{\textcolor{orange}{boda-boda}} per muoversi a \textbf{\textcolor{purple}{Goma}}. Il \textbf{\textcolor{pink}{prezzo normale}} è di circa \textbf{\textcolor{blue}{500 franchi congolesi}} per una corsa breve. \\
    Chinese: & 你可以乘坐 \textbf{\textcolor{orange}{boda-boda}}游览戈马\ (\textbf{\textcolor{purple}{gema}})。短途车程的正常价格是 500 刚果法郎\ (\textbf{\textcolor{blue}{gangguo falang}})。 \\
  \end{tabularx}

  \vspace{3pt}
\end{mdframed}
\end{CJK*}

\caption{%
  Example parallel utterances from FLEURS, where highlighted items share
  pronunciation similarities that spoken translation retrieval could rely on
  instead of semantic cues. Colors indicate phonetically similar items.%
}
\label{tab:pron-shortcuts}
\end{table}

\begin{enumerate}
    \item[] \textbf{Cross-lingual speech retrieval is possible without pronunciation cues.} We propose a challenge set devoid of \textit{pronunciation}-level cues, where we show cross-lingual retrieval accuracy can remain strong particularly in the final layers, despite noticeable performance drops.\footnote{We release the dataset here: \url{https://anonymous.4open.science/r/pronunciation-challenge-set-6214/}}
    \item[] \textbf{Speech translation objective drives semantics in Whisper-style models.} Importantly, we show that the spoken translation retrieval capabilities in Whisper-style speech encoders are driven strongly by the existence of a speech translation objective, rather than arising from next-token prediction objective in multilingual ASR.

    \item[] \textbf{Early exiting the encoder generalizes better to related languages.} Leveraging our insights, we early exit the ASR encoder to induce representations less tied to language-specific semantics. Applying such an early exiting approach to ASR on seven low-resource languages, we observe consistent improvements in error rates over using the final encoder layer on 5 out of 7 low-resource languages examined.
\end{enumerate}

\section{Related Work}
In this section, we review prior work that analyze cross-lingual alignment in audio representations, along with work that looks into early exiting neural models.

\begin{figure*}
    \centering
        \includegraphics[width=0.65\linewidth]{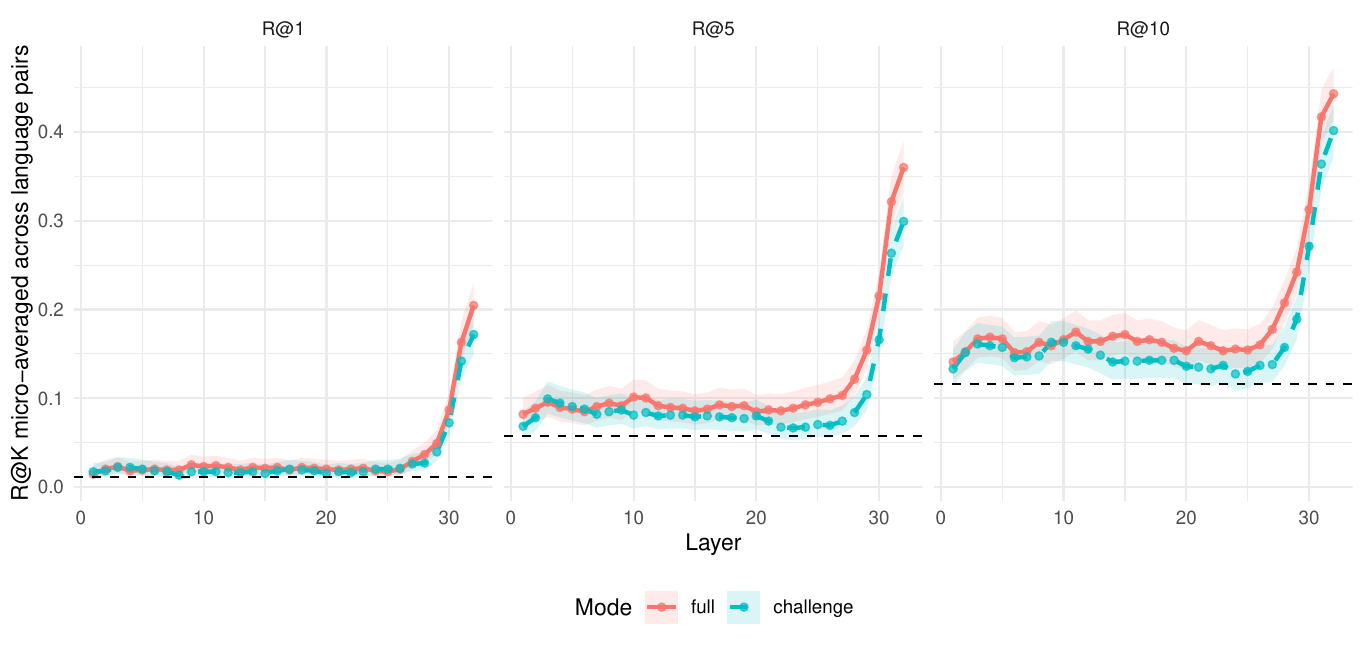}
    \caption{Spoken translation retrieval results in Whisper-large-v2. Plot shows R@1, R@5, and R@10, micro-averaged across language pairs. Shaded regions indicate 95\% Wilson confidence intervals. Dashed line is a random baseline. We observe that even after filtering out potential pronunciation shortcuts, semantic-based retrieval remains strongly above chance towards the later layers for all R@K values.}
    \label{fig:whisper_retrieval}
\end{figure*}

\subsection{Cross-Lingual Alignment in Speech}

Cross-lingual alignment is well established in multilingual text encoders such as mBERT \cite{devlin-etal-2019-bert} and XLM \cite{lample2019crosslinguallanguagemodelpretraining}. For speech models, however, it remains unclear to what extent semantic alignment emerges, especially when training is dominated by multilingual ASR, where next-token prediction is tightly coupled to the input signal.
Recent work on  probing speech representations suggests that self-supervised models are predominantly phonetic \cite{choi24b_interspeech}, though such analyses often rely on short word-level inputs. In contrast, models trained with speech translation objectives show strong spoken translation retrieval performance \cite{abdullah24_interspeech, ma2025crosslingualtransferlearningspeech}, suggesting the emergence of a shared semantic subspace. Further, \cite{ogunremi2025transcribe} observe that Whisper-style encoders exhibit an English-centered interlingua, hypothesized to arise from the speech translation objective.
However, retrieval-based evaluations may be inflated by pronunciation-level cues such as cognates and proper nouns, and it remains unclear whether semantic alignment is primarily driven by speech translation supervision or can emerge from multilingual ASR alone. We address both questions by (i) constructing a pronunciation-controlled challenge set across typologically distant languages, and (ii) comparing Whisper-style encoders trained with and without speech translation objectives.

\subsection{Early Exiting}
Prior work has shown that representations at earlier layers of a neural network are often already adequate for making a correct prediction \cite{kaya2019shallowdeepnetworksunderstandingmitigating}, which allows for early exiting strategies that minimize compute \cite{schuster2022confidentadaptivelanguagemodeling} and mitigate the influence of false demonstrations \cite{halawi2024overthinkingtruthunderstandinglanguage}. Recent work in mechanistic interpretability \cite{rai2025practicalreviewmechanisticinterpretability} extend this line of research by using early exiting as a way to understand model internals. For instance, the logit lens method \cite{nostalgebraist2020logitlens} interprets the incremental layer updates of large language models by projecting intermediate hidden states into the vocabulary space. However, the work above has mainly been applied to decoder-based text models. \cite{langedijk-etal-2024-decoderlens} adapt the method to encoder-decoder architectures and also apply it to Whisper. However, their work aims to identify at what encoder layer stable predictions emerge, and focuses on the tasks Whisper was explicitly trained on. We extend this line of work and apply early exiting to strengthen model generalization,  providing concrete applications as to how it can aid low-resource speech recognition.

\section{Experimental Setup}
In this section, we describe the methods we employ, along with our dataset construction procedure.

\begin{figure*}
    \centering
        \includegraphics[width=0.65\linewidth]{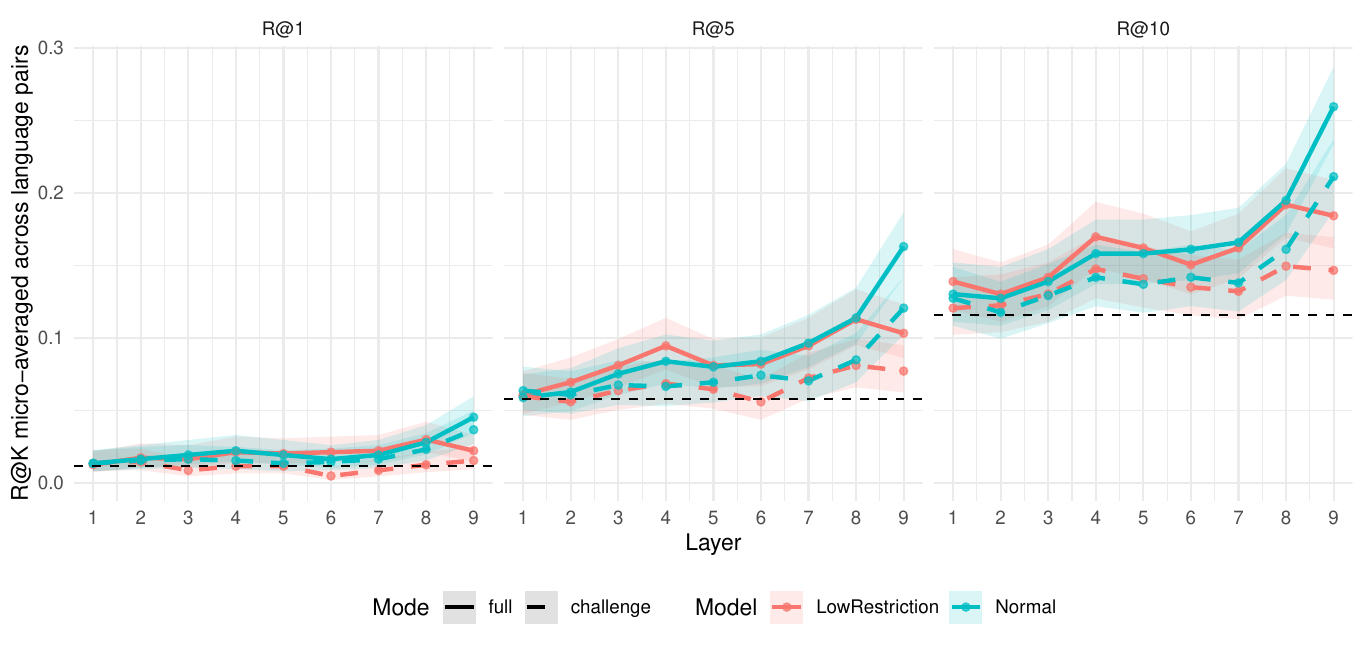}
    \caption{Spoken translation retrieval results in models with and without speech translation objective. Plots show R@1, R@5, and R@10, micro-averaged across language pairs. Shaded regions indicate 95\% Wilson confidence intervals. Dashed line is a random baseline. The model with an additional speech translation objective (Normal) shows stronger retrieval accuracy than the model without such an objective in the final layer, although the latter model (LowRestriction) also exhibits non-trivial retrieval accuracy at R@10.}
    \label{fig:owsm_st}
\end{figure*}

\begin{table}[H]
\centering
\footnotesize
\begin{tabular}{lrrr}
\toprule
\textbf{Langs} & \textbf{Original Test Set}& \textbf{Challenge Set} & \textbf{Sampled Test Set}\\
\midrule
eng–zho   & 427 &  100 &  100 \\
fra–zho   & 427 &  108 &  108 \\
deu–zho   & 427 &  94 &  94 \\
eng–jpn   & 427 &   77 &   77 \\
fra–jpn   & 427 &   72 &   72 \\
deu–jpn   & 427 &   67  &   67 \\
\bottomrule
\end{tabular}
\caption{Statistics for the FLEURS dataset employed in our study (original test set size, our challenge set size, and sampled test set size, where we perform sampling on the original test set to ensure size comparability with the challenge set).}
\label{tab:dataset-stats}
\end{table}

\subsection{Model}
In our experiments, we follow \cite{ma2025crosslingualtransferlearningspeech} in examining Whisper \cite{radford2022whisper} as a case study. Whisper is an encoder-decoder Transformer model trained on the tasks of language identification, speech recognition, and X $\rightarrow$ English speech translation. It is trained on 680,000 hours of data, where 10\% is for the task of X $\rightarrow$ English translation. Audio is first converted to a mel spectrogram, then passed through convolutional layers to extract features that are then passed to the Transformer's blocks.  In this paper, we use the whisper-large-v2 model, which has been shown to exhibit stronger cross-lingual alignment than later variants \cite{ma2025crosslingualtransferlearningspeech}. In addition, to observe the effect of the speech translation objective on encoder representations, we employ Open Whisper-style Speech Models (OWSM), which provide a series of open-source, Whisper-style encoder-decoder models. OWSMs provide an ideal testing ground for this hypothesis due to the open source nature of its training data; their adherence to Whisper-style training specifications; and they provide models of various sizes and setups. OWSM includes two models of comparable size and architecture, OWSM v3.1 Small and Small Low-Restriction, of which the latter is not trained on speech translation. As such, we repeat our spoken translation retrieval experiments on the two models to gain insights on the impact of the speech translation objective.

\subsection{Datasets}

For our speech retrieval experiments, we follow \cite{ma2025crosslingualtransferlearningspeech} in employing FLEURS \cite{fleurs}, a multilingual parallel speech dataset. They build their test set by combining the dev and test sets of English, French, German, Mandarin and Japanese, where only utterances common to all five languages are kept. This is followed by a deduplication step to remove utterances with the same content spoken by different speakers. We build our challenge set on top of this procedure. We employ the spaCy \cite{honnibal2020spacy} pipeline to filter out utterances containing proper nouns. For pairs with Japanese, we further remove samples containing katakana script, which are largely reserved for loanwords in Japanese. We pair together typologically and phylogenetically distant languages to avoid the influence of cognates between related varieties, resulting in six language pairs (\autoref{tab:dataset-stats}). Proficient in-house speakers of the five languages then manually inspect the resulting dataset to confirm their quality, which further removes utterances with similar sounding measurement units (e.g.\ GHz). As cross-lingual retrieval is direction-dependent, we perform our experiments on both src-trg and trg-src retrieval directions. To keep language pair proportions comparable, we sample the full set to contain the same language pair proportions as the challenge set.%

For our early exiting experiments, we use the FLEURS test sets of the same 7 low-resource languages examined in \cite{ma2025crosslingualtransferlearningspeech}, which include Cebuano (ceb), Irish Gaelic (ga), Javanese (jv), Asturian (ast), Kabuverdianu (kea), Sorani Kurdish (ckb), and Kyrgyz (ky).

\section{Methodology}

\begin{table*}[tbp]
\centering
\begin{CJK*}{UTF8}{gbsn}
\begin{mdframed}[style=GreenBox]
\footnotesize
\setlength{\tabcolsep}{3pt}

\begin{tabular}{l*{19}{c}}
\textbf{Last}:
& Ubelak & lorok & bubuk & garing
& \textcolor{eqA}{itu}
& bebarengan
& \textcolor{eqB}{dengan}
& \textcolor{eqC}{banjir.}
& Kandai & tangan
& \textcolor{eqD}{telah}
& 
& sengerasik,
& remes
& \textcolor{eqE}{itu}
& \textcolor{eqF}{sudah}
& tebal.
& \phantom{bal}
& \phantom{bal}
\\

\textbf{Best}:
& Uplok & lorok & bubuk & garing
& \textcolor{eqA}{iku}
& bebarengan
& \textcolor{eqB}{lant}
& \textcolor{eqC}{banjur.}
& Kanti & tangan
& \textcolor{eqD}{telas}
& yang & resik,
& remes
& \textcolor{eqE}{iku}
& \textcolor{eqF}{dati}
& bal.
& \phantom{bal}
& \phantom{bal}
\\

\textbf{Gold}:
& ublek & loro & bubuk & garing
& \textcolor{eqA}{iku}
& bebarengan
& \textcolor{eqB}{lan}
& \textcolor{eqC}{banjur}
& kanthi & tangan
& \textcolor{eqD}{teles}
& sing & resik
& remes
& \textcolor{eqE}{iku}
& \textcolor{eqF}{dadi}
& bal
& \phantom{bal}
& \phantom{bal}
\\

\end{tabular}

\vspace{3pt}
\end{mdframed}
\end{CJK*}
\caption{Example prediction of early exiting Whisper-large-v2 on Javanese, a language not explicitly supported by the model. Colored words indicate comparability. In comparison against the reference, standard inference (layer 32) heavily uses Indonesian words, whereas early exiting (layer 29) results in transcriptions more phonetically faithful to Javanese, resulting also in lower error rates.}
\label{tab:early-exit-example}
\end{table*}

\begin{table}[t]
  \centering
  \footnotesize
  \setlength{\tabcolsep}{3pt}
  \begin{tabular}{lccrccr}
    \toprule
    & \multicolumn{3}{c}{\textbf{CER (\%)}}
    & \multicolumn{3}{c}{\textbf{WER (\%)}} \\
    \cmidrule(lr){2-4} \cmidrule(lr){5-7}
    \textbf{Lang}
      & \textbf{Best}
      & \textbf{L32}
      & \textbf{$\Delta$}
      & \textbf{Best}
      & \textbf{L32}
      & \textbf{$\Delta$} \\
    \midrule
    ky  & \textbf{100.5} (24) & 138.2 & 37.7
        & \textbf{106.6} (24) & 194.3 & 87.7 \\
    ga  & \textbf{79.2} (31)  & 85.5  & 6.3
        & \textbf{101.3} (24) & 121.8 & 20.5 \\
    jv  & \textbf{36.1} (29)  & 41.3  & 5.2
        & \textbf{87.3} (29)  & 90.8  & 3.5 \\
    ceb & \textbf{17.6} (29)  & 19.2  & 1.7
        & \textbf{52.9} (29)  & 54.4  & 1.5 \\
    kea & \textbf{34.4} (29)  & 35.1  & 0.8
        & \textbf{91.6} (29)  & 92.6  & 1.0 \\
    ckb & \textbf{57.2} (32)  & \textbf{57.2} & 0.0
        & \textbf{102.6} (25) & 122.4 & 19.8 \\
    ast & \textbf{16.4} (32)  & \textbf{16.4}  & 0.0
        & 59.6 (29)  & \textbf{58.7} &  -0.9 \\
    \bottomrule
  \end{tabular}
  \caption{Early exiting results on FLEURS test set of low-resource languages. \textbf{Best} indicates performance and best layer
selected (in parenthesis) on the dev set; \textbf{L32} is the final encoder layer performance. $\Delta$ shows performance improvement.}
  \label{tab:cer-wer}
\end{table}

\subsection{Cross-Lingual Speech Retrieval}
To quantify cross-lingual alignment in the encoder of Whisper-style models, we follow prior work in employing translation retrieval as a proxy task. \cite{ma2025crosslingualtransferlearningspeech} propose SeqSim to quantify the similarity between two sequences of audio embeddings $X = \{ \mathbf{x}_1, \dots, \mathbf{x}_m \}$ and $Y = \{ \mathbf{y}_1, \dots, \mathbf{y}_n \}$ by measuring how well each frame in one sequence aligns with the most similar frame in the other, where the process is repeated in both directions. The benefit of this measure is that it is simple and allows frames in two audio embeddings to match one another regardless of their position, where mean pooling frame representations in contrast risks cramming too much information due to the large number of frames. 
Formally, SeqSim is defined as:

\begin{equation}
\scalebox{0.85}{$
\begin{aligned}
\text{Re\textsubscript{seq}}(X, Y) &= \frac{1}{|X|} \sum_{\mathbf{x} \in X} \max_{\mathbf{y} \in Y} \mathbf{x}^\top \mathbf{y}, \\
\text{Pr\textsubscript{seq}}(X, Y) &= \frac{1}{|Y|} \sum_{\mathbf{y} \in Y} \max_{\mathbf{x} \in X} \mathbf{x}^\top \mathbf{y}, \\
\text{SeqSim}(X, Y) &= 2 \cdot \frac{\text{Pr\textsubscript{seq}} \cdot \text{Re\textsubscript{seq}}}{\text{Pr\textsubscript{seq}} + \text{Re\textsubscript{seq}}}.
\end{aligned}
$}
\end{equation}

 \cite{ma2025crosslingualtransferlearningspeech} show SeqSim outperforms mean pooling and dynamic time warping for spoken translation retrieval. We thus leverage this metric in our work.

\subsection{DecoderLens}
In this work, we follow \cite{nostalgebraist2020logitlens} in viewing the layers of a transformer as performing incremental updates to latent predictions of the next token. This assumption implies that the hidden states can be decoded to gain insights as to how the input is being processed in said layer. This method has gained widespread usage in text-based decoder-only language models, where it is used to illustrate what the model captures in terms of vocabulary space at each internal layer \cite{belrose2023elicitinglatentpredictionstransformers,din2024jumpconclusionsshortcuttingtransformers}. \cite{langedijk-etal-2024-decoderlens} extend this method to encoder-decoder models, where the decoder is used as a probe over intermediate encoder representations. Concretely, let $\mathcal{M}$ be an encoder-decoder model with $n$ encoder layers. At inference time, the model typically produces an output by passing the final encoder layer's representation to the decoder (after a non-linear transformation such as layer normalization), i.e.,
\begin{equation}
\mathcal{M}(\mathbf{w}) = \mathrm{Dec}\big(f(\mathrm{Enc}(\mathbf{w})_n)\big),
\end{equation}
where $\mathrm{Enc}(\mathbf{w})_n$ denotes the hidden states at the final encoder layer and $f$ captures the non-linear operation applied prior to decoding. DecoderLens modifies this procedure by replacing the final-layer encoder representation with that of an intermediate layer $i$. Formally, for encoder layer $i \in \{1, \dots, n\}$, DecoderLens computes:
\begin{equation}
\mathrm{DecoderLens}(\mathbf{w}, i) = \mathrm{Dec}\big(f(\mathrm{Enc}(\mathbf{w})_i)\big).
\end{equation}

Building on the findings that the final encoder layer's representations are more semantically aligned, we early exit the encoder to induce representations less tied to language-specific semantics, which we hypothesize to generalize better to speech recognition on low-resource languages by weakening such target-language specialization. 
For our early exiting experiments on the low-resource languages examined in \cite{ma2025crosslingualtransferlearningspeech}, we select the best performing layer based on WER and CER on the dev set of FLEURS, and use that layer to perform inference on the FLEURS test sets of the respective language. As our goal is only to reduce the influence of language-specific semantic information, we perform this experiment on the 8 encoder layers of 24 to 32.

\section{Results}

In this section, we detail our results on spoken translation retrieval and early exiting the encoder.

\subsection{Cross-lingual speech retrieval is possible without pronunciation cues.} First, we compare the results on the full data to our challenge set with the Whisper encoder. \autoref{fig:whisper_retrieval} shows our spoken translation retrieval results in Recall@K across different language pairs using the Whisper-large-v2 encoder. For all R@K setups, performance is significantly above the random baseline in the final layers. This indicates that although phonetic cues are indeed useful for retrieval, the models examined are able to rely mostly on semantic cues. 

\subsection{Spoken translation retrieval in Whisper-style models is driven primarily by the speech translation objective.} 
\autoref{fig:owsm_st} shows our experiment in comparing OWSM Small and OWSM Small Low-restriction, where the latter differs minimally from the former in lacking the speech translation objective. We observe that on both the full and the challenge set, the model not trained on speech translation (OWSM Small Low-restriction) exhibits weaker retrieval capabilities in comparison with the model with such an objective (OWSM Small). For OWSM Small, the gap is particularly pronounced for the final layers, where we observe stronger performance under all retrieval settings. This suggests that speech translation is indeed a strong contributor for semantic cross-lingual alignment capabilities in speech models, which nevertheless can still arise to a much less degree when trained only on multilingual ASR. %

\subsection{Encoder representations in earlier layers generalize better to low-resource languages.}

Next we evaluate early exiting, to measure to what degree our findings may impact automatic speech recognition quality by leveraging representations in pre-final layers.
\autoref{tab:cer-wer} shows our results in using encoder representations in earlier layers to perform speech recognition on low-resource languages. We observe that interestingly 5 out of 7 languages examined show improved CER and WER using some earlier encoder layer's representation, implying that earlier encoder layers may generalize better to low-resource languages not explicitly supported by Whisper. \autoref{tab:early-exit-example} provides an example of our early exited prediction on Javanese, where we observe the final layer to produce transcriptions with significantly more Indonesian words than the early exited transcriptions, which are more phonetically faithful to the Javanese gold transcriptions.

\section{Discussion and Conclusion}
In this work, we revisit the question of cross-lingual alignment in Whisper-style speech foundation models. Through a series of controlled experiments, %
we demonstrate that spoken translation retrieval remains possible in Whisper-style speech encoders even without %
phonetic cues—such as cognates and proper nouns. %
Importantly, we demonstrate that spoken translation retrieval %
is strongly driven by the additional speech translation pre-training objective, although it also arises weakly through speech recognition as an objective alone, which underscores the role of supervised multilingual multi-task signals in pre-training to aid shaping useful semantic representations. Finally, by early exiting the encoder to induce representations less tied to language-specific semantics, we observe that using earlier encoder layer representations results in performance gains in speech recognition on low-resource languages, demonstrating the downstream applicability of our findings. As Whisper-style phone recognition models have been shown to perform weaker on sociolinguistic variation and unseen languages \cite{li2026powsmphoneticopenwhisperstyle}, future work can look into whether our early-exiting findings extend also to such phone recognition models.

\section{Generative AI Use Disclosure}
The authors acknowledge the usage of ChatGPT as an assistant tool in part of the source code’s development and in enhancing the coherence of parts of the manuscript.

\bibliographystyle{IEEEtran}
\bibliography{mybib}

\end{document}